\documentclass[letterpaper]{article} 
\usepackage{aaai23} 
\usepackage{times} 
\usepackage{helvet} 
\usepackage{courier} 
\usepackage[hyphens]{url} 
\usepackage{graphicx} 
\usepackage{natbib} 
\usepackage{caption} 
\usepackage{algorithm}
\usepackage{algorithmic}

\usepackage{amsfonts,amssymb}
\usepackage{amsmath}
\usepackage{multirow}
\usepackage{booktabs}
\usepackage{microtype}
\usepackage[american]{babel}
\usepackage{bm}

\usepackage{epstopdf}
\usepackage{subfigure}
\usepackage{xcolor}

\pdfoutput=1

\usepackage{newfloat}
\usepackage{listings}
\DeclareCaptionStyle{ruled}{labelfont = normalfont,labelsep = colon,strut = off} 
\floatstyle{ruled}
\newfloat{listing}{tb}{lst}{}
\floatname{listing}{Listing}
\title{Refined Semantic Enhancement towards Frequency Diffusion for Video Captioning}
\author {
 Xian Zhong,\textsuperscript{\rm 1}
 Zipeng Li,\textsuperscript{\rm 1}
 Shuqin Chen,\textsuperscript{\rm 2,}\footnote{Corresponding authors.}
 Kui Jiang,\textsuperscript{\rm 3,$\ast$}
 Chen Chen,\textsuperscript{\rm 4}
 and Mang Ye \textsuperscript{\rm 3}
}
\affiliations {
 \textsuperscript{\rm 1} School of Computer Science and Artificial Intelligence, Wuhan University of Technology\\
 \textsuperscript{\rm 2} College of Computer, Hubei University of Education\\
 \textsuperscript{\rm 3} School of Computer Science, Wuhan University\\
 \textsuperscript{\rm 4} Center for Research in Computer Vision, University of Central Florida\\
 \{zhongx, lizipeng, csqcwx0801\}@whut.edu.cn, kuijiang@whu.edu.cn, chen.chen@crcv.ucf.edu, mangye16@gmail.com
}

\usepackage{bibentry}

\begin{document}

\maketitle


\begin{abstract}
Video captioning aims to generate natural language sentences that describe the given video accurately. Existing methods obtain favorable generation by exploring richer visual representations in encode phase or improving the decoding ability. However, the long-tailed problem hinders these attempts at low-frequency tokens, which rarely occur but carry critical semantics, playing a vital role in the detailed generation. 
In this paper, we introduce a novel Refined Semantic enhancement method towards Frequency Diffusion (RSFD), a captioning model that constantly perceives the linguistic representation of the infrequent tokens. Concretely, a Frequency-Aware Diffusion (FAD) module is proposed to comprehend the semantics of low-frequency tokens to break through generation limitations. 
In this way, the caption is refined by promoting the absorption of tokens with insufficient occurrence. Based on FAD, we design a Divergent Semantic Supervisor (DSS) module to compensate for the information loss of high-frequency tokens brought by the diffusion process, where the semantics of low-frequency tokens is further emphasized to alleviate the long-tailed problem. 
Extensive experiments indicate that RSFD outperforms the state-of-the-art methods on two benchmark datasets, \textit{i.e.}, MSR-VTT and MSVD, demonstrate that the enhancement of low-frequency tokens semantics can obtain a competitive generation effect.
Code is available at \textcolor{magenta}{\url{https://github.com/lzp870/RSFD}}.
\end{abstract}

\section{Introduction}
Video captioning is the task of understanding video content and describing it accurately. It has considerable application value in social networks, human-computer interaction, and other fields.
Despite recent progress in this field, it remains a challenging task as existing models prefer to generate commonly-used words disregarding many infrequent tokens that carry critical semantics of video content, limiting the refined semantic generation in the testing phase.

\begin{figure}
	\centering
	\includegraphics[width = \columnwidth]{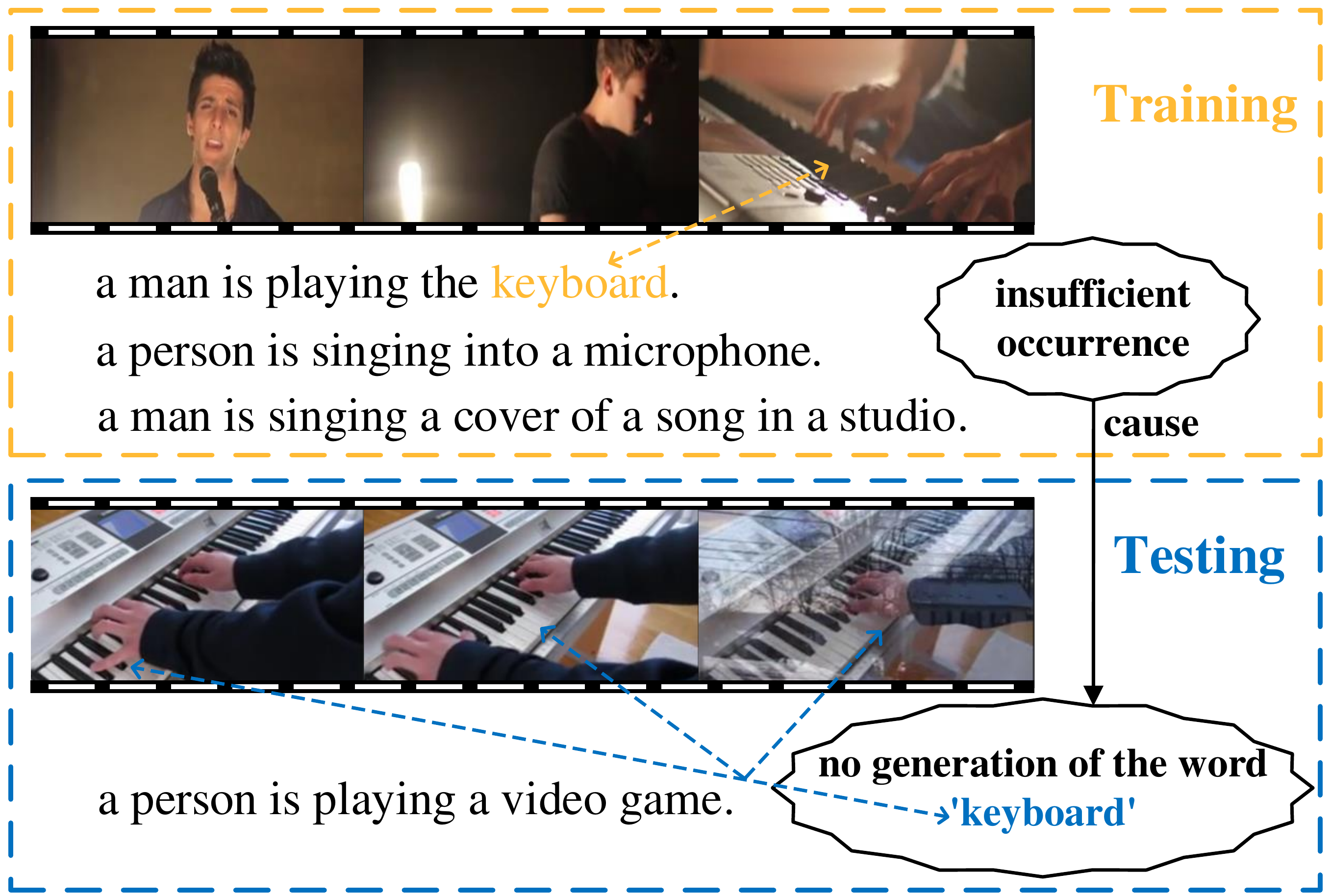} 
	\caption{Illustration of existing models limitations: the insufficient occurrence of low-frequency tokens affects the information extraction during training, thus failing to generate refined low-frequency words in the testing phase.}
	\label{fig1}
\end{figure}

\begin{figure*}
	\centering
	\includegraphics[width = \textwidth]{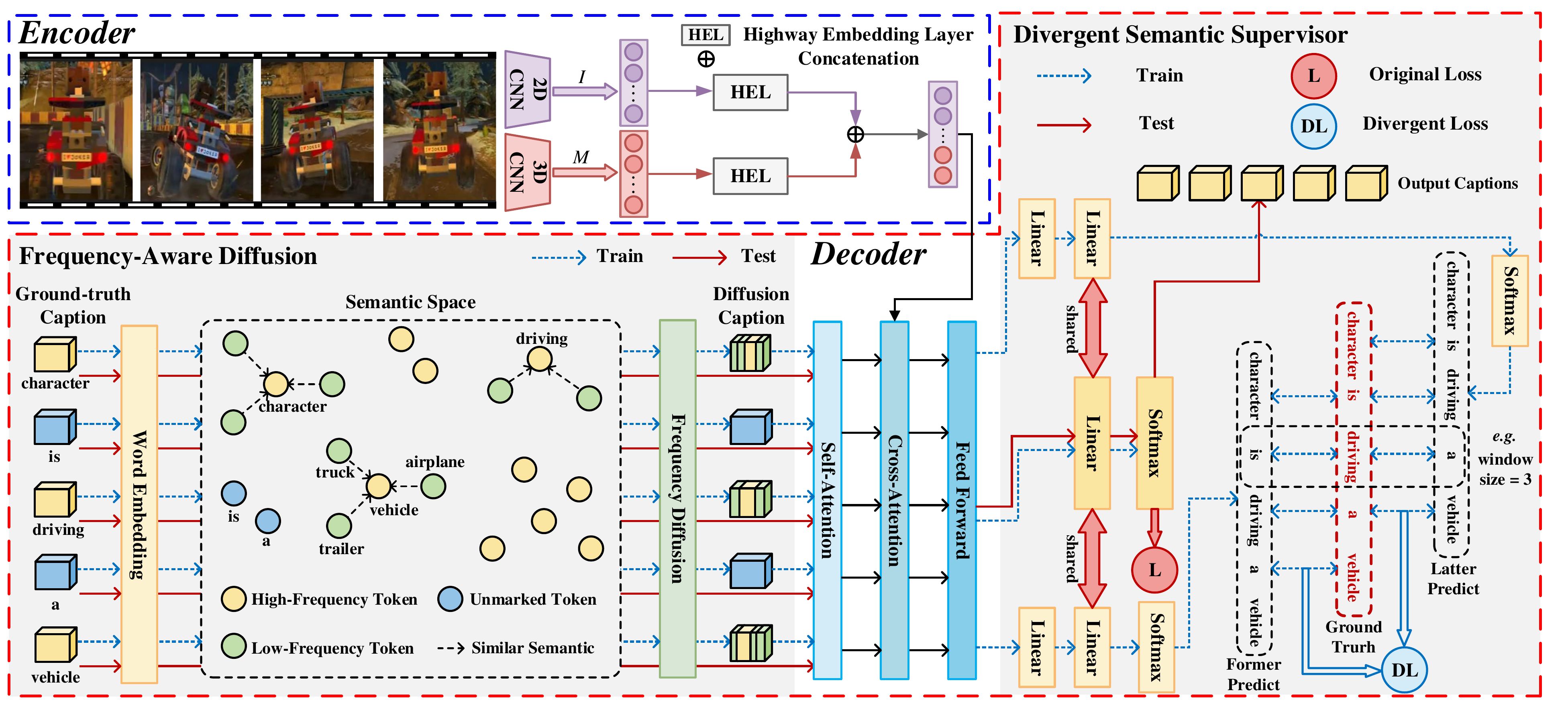} 
	\caption{Overview of the proposed RSFD architecture. It mainly consists of the encoder in the top-left box and the decoder with FAD and DSS modules in another box. In the training phase, FAD promotes the model comprehending the refined information by mapping the ground-truth caption to the semantic space and fusing it in frequency diffusion. DSS supervises the central word to obtain its distinctive semantics. In the testing phase, only the transformer-based parts are implemented for sentence generation.}
	\label{fig2}
\end{figure*}

The existing methods generally adopt the encoder-decoder framework~\cite{VenugopalanXDRM15}, where the encoder generates visual representation by receiving a set of consecutive frames as input, and the decoder generates captions via recurrent neural networks (RNNs) or Transformer~\cite{HoriHLZHHMS17}. 
Some efforts~\cite{HuJLN0W22, HuJL0J022, JiaZYLH22, LiaoXWCSYL22, XuLWHT22} are developed to explore richer visual features.
Prior works~\cite{ZhangP19a, PanCHLGAN20, ZhangSY0WHZ20} enhance the spatio-temporal representations between objects, while others~\cite{JinHCLZ20, GaoHSSGJW22} tend to improve the architecture of the decoder to obtain better linguistic representations.
However, they either focus on the enhancement of visual representations or the generation ability of the model, ignoring the tokens with insufficient occurrence in the training phase. Many low-frequency tokens, as a critical factor to prove caption performance, have not received adequate attention, thus challenging for the model to express similar information in other videos in the testing phase like the instance in Fig.~\ref{fig1}. Concretely, a man is playing the keyboard and the model has partially generated ``a person is playing a video game''. The result commonly occurs when the low-frequency token ``keyboard'' is of insufficient occurrence in the given video.

To deal with the challenges of token imbalance, the approach~\cite{WuSCLNMKCGMKSJL16} in neural machine translation splits words into more fine-grained units. However, the token imbalance phenomenon is not fundamentally eliminated. 
BMI~\cite{XuLMZX020} has been dedicated to assigning adaptive weights to target tokens in light of the token frequency. Although they increase the exposure of low-frequency tokens during training, they potentially damage the information of high-frequency ones. 
In video captioning, few researchers focus on the learning effect of low-frequency tokens. 
ORG-TRL~\cite{ZhangSY0WHZ20} introduces an external language model to extend the low-frequency tokens, which depends heavily on external knowledge.
Taking the above attempts as a reference, the motivation of our design can be depicted as follows:
1) With the help of high-frequency tokens more times trained by the model, low-frequency tokens entrust their semantics to high-frequency ones obtaining adequate exposure to alleviate the effect of token imbalance.
2) Instruction from multiple teachers to high-frequency tokens boosts specific semantics suitable for them as compensation due to information loss by the previous stage, and low-frequency tokens further enhance their generation.

In this work, we propose a novel method named Refined Semantic enhancement towards Frequency Diffusion (RSFD) for video captioning to address the mentioned issue. In RSFD, unlike previous effort~\cite{ZhangSY0WHZ20} that with the help of external modules, we propose a module named Frequency-Aware Diffusion (FAD) to explore the semantics of low-frequency tokens inside our model.
Firstly, in light of the occurrence numbers of each token in the whole corpus and in each video they appear, we split them into three categories (high-frequency, low-frequency, and unmarked).
Secondly, we exploit the diffusion process to comprehend low-frequency tokens by integrating them with high-frequency ones. 
With the assistance of the diffusion of low-frequency tokens, the model develops the ability to absorb the semantics of low-frequency tokens. 
Afterward, to alleviate the semantic loss of high-frequency tokens (Pure high-frequency words can be entirely trained, whereas they carry low-frequency semantics that disturbs the model's learning of them) by the diffusion process and further excavate the low-frequency information, we further design a Divergent Semantic Supervisor (DSS) module to exploit adjacency words to supervise the central word. 
With the adjacent supervisory information, the central high-frequency token can obtain the semantics from multiple teachers adapted to itself for compensation, and the central low-frequency one can further promote its generation. 
We concentrate on the representation of low-frequency tokens, which helps the model capture information about relatively rare but vital words in the testing phase. The improvement of low-frequency tokens enables the generation caption to have a refined description.

Our main contributions are summarized threefold:

\begin{itemize}
\item We put the emphasis on token imbalance that has not been focused seriously in video captioning and dedicate to concentrating on infrequent but vital tokens to enhance the refined semantics of generated captions. 
\item We devise a unique module, termed Frequency-Aware Diffusion (FAD), in the form of diffusion to learn tokens with insufficient occurrence that are most likely to carry critical semantics. To the best of our knowledge, this is the first time that the frequency of tokens is attached to great importance in video captioning.
\item We further design a Divergent Semantic Supervisor (DSS) module as a constraint for balancing different frequent tokens by leveraging distinctive semantics adapted to the token itself, and quantitatively validating its effectiveness.

\end{itemize}

\section{Related Works}
\subsection{RNN- and Transformer-Based Methods}
With the development of deep learning, researchers extract video features through deep convolution neural networks and combine them with RNNs or Transformer~\cite{HoriHLZHHMS17} to make the generated words more accurate. 
Early in~\cite{VenugopalanXDRM15}, the encoder-decoder framework is applied first to the captioning model, where visual features are leveraged by performing mean-pooling over each frame, and long short-term memory (LSTM)~\cite{HochreiterS97} is adopted to generate sentences. 
Benefiting from parallel computing ability and scalability of Transformer~\cite{Transformer}, it is applied to many multi-modal tasks. Inspired by it, Transformer has been introduced for video captioning~\cite{ChenLZH18}. 
SBAT~\cite{JinHCLZ20} is intended to reduce the redundancy of consecutive frames. {D}$^2$~\cite{GaoHSSGJW22} utilizes the syntactic prior to measuring the contribution of visual features and syntactic prior information for each word. 

\subsection{Diffusion Model}
Some efforts~\cite{LiZCDZY22,YeLDSSH22} have been conducted on data noise.
In recent years, Diffusion model~\cite{Sohl-DicksteinW15} has made remarkable progress. 
DDPM~\cite{HoJA20} develops the denoising ability by artificially adding noise in the diffusion process and matching each small step of the inverse process corresponding to the forward process. 
TimeGrad~\cite{RasulSSV21} integrates the diffusion model with the autoregressive model by sampling from the data distribution at each time step by estimating its gradient.
The diffusion approaches~\cite{LiaoCXWLS22,Zhu2022FineGF} have obtained significant progress in computer vision.
Inspired by the potential of learning noise by noise addition that is just aiming at learning the noise purely, RSFD regards the information on low-frequency tokens as noise and adds it to high-frequency tokens, developing the model's ability to comprehend low-frequency tokens.

\subsection{Translation Methods with Frequency Modeling}
In neural machine translation, the token imbalance data drive the model preferentially generates high-frequency words while ignoring the rarely-used words. GNMT~\cite{WuSCLNMKCGMKSJL16} proposes to divide words into a limited set of common sub-word units to handle the rare words naturally.
BMI~\cite{XuLMZX020} assigns an adaptive weight to promote token-level adaptive training. 
In video captioning, the large-scale training corpus demonstrates the phenomenon of the imbalance occurrence of tokens. 
External language model~\cite{ZhangSY0WHZ20} is introduced to extend the ground-truth tokens to deal with the token imbalance problem, which heavily depends on external knowledge. In contrast, RSFD highlight relatively rare words that are most likely refined semantics in the sentence, alleviating token imbalance inside our model.

\section{Proposed Method}
We devise our Refined Semantic enhancement towards Frequency Diffusion (RSFD) for video captioning. As shown in Fig.~\ref{fig2}, our overall framework follows the encoder-decoder structure. 
During the training process, Frequency-Aware Diffusion (FAD) encourages the model to add low-frequency token noise to learn its semantics. Then the diffusion features of tokens are fused with the corresponding visual features according to the cross-attention mechanism. 
At the head of the decoder, Divergent Semantic Supervisor (DSS) obtains distinctive semantic features by updating the gradient that adapts to the token itself. In the testing phase, only the original Transformer architecture is retained to generate captions.

\subsection{Encoder-Decoder Framework}

\subsubsection{Encoder.}
Image and motion features are processed by two separate encoders, which give a set of consecutive video clips of length $K$. Here we use $V \in \mathbb{R}^{K \times d_v}$ to represent the features of these two types, where $d_v$ denotes dimension of each feature. 
We feed them into highway embedding layer (HEL) to obtain representations $R \in \mathbb{R}^{K \times d_h}$, \textit{i.e.}, $R = f_\text{HEL}(V)$, where $d_h$ is output dimension of HEL, and $f_\text{HEL}$ can be trained directly through simple gradient descent~\cite{SrivastavaGS15}, thus it can be formulated as:
\begin{small}
\begin{equation}
	f_\text{HEL}(V) = \text{BN} \left(T \left(V\right) \circ H \left(V\right) + \left(1-T \left(V\right)\right) \circ P \left(V\right)\right),
\end{equation}
\end{small}
where $H(V) = V W_\text{eh}$, $P(V) = \tanh(H(V) W_\text{ep})$, and $T(V) = \phi(H(V) W_\text{et})$.
BN denotes batch normalization operation~\cite{IoffeS15}, $\circ$ is the element-wise product, and $\phi$ is the sigmoid function. $W_\text{eh} \in \mathbb{R}^{d_v \times d_h}$, and $\{W_\text{ep}, W_\text{et}\} \in \mathbb{R}^{d_h \times d_h}$. The multi-modalities, \textit{e.g.}, image and motion features are applied concatenation to obtain ${R}\in\mathbb{R}^{2K \times d_h}$ to feed into the decoder.

\subsubsection{Decoder.}
The decoder block consists of self-attention, cross-attention, and feed-forward layer. The self-attention layer is formulated as:
\begin{small}
\begin{equation}
\begin{aligned}
	\text{SelfAtt} \left(E_{<t}\right) & = \text{MultiHead} \left(E_{<t},E_{<t},E_{<t}\right)\\
	& = \text{Concat} \left(\text{head}_{1:h}\right)W_h,
\end{aligned}
\end{equation}
\end{small}
where $E_{<t} \in \mathbb{R}^{T \times d_h}$ denotes the feature vector of embedded captions, $T$ represents the length of caption. Concat denotes concatenation operation, and $h$ denotes the number of multiple attention heads. $W_h \in \mathbb{R}^{d_h \times d_h}$ is a trainable parameter. 
Residual connection and layer normalization are adopted after the self-attention layer:
\begin{small}
\begin{equation}
	E^{'}_{<t} = \text{LayerNorm} \left(E_{<t} + \text{SelfAtt} \left(E_{<t}\right)\right),
\end{equation}
\end{small}
where LayerNorm smooths the size relationship between different samples. 

So far, the self-attention process has been completed, and the result $E^{'}_{<t}$ is employed for cross-attention as a query:
\begin{small}
\begin{equation}
	D_{<t} = \text{LayerNorm} \left(E^{'}_{<t} + \text{MultiHead} \left(E^{'}_{<t}, R, R\right)\right),
\end{equation}
\end{small}
where $D_{<t}$ denotes the sentence feature of cross-attention output. We adopt a feed-forward network (FFN) that employs non-linear transformation,
and apply LayerNorm to calculate the probability distribution of words by Softmax:
\begin{small}
\begin{equation}
	D_t = \text{LayerNorm} \left(D_{<t} + \text{FFN} \left(D_{<t}\right)\right),
\label{eq5}
\end{equation}
\begin{equation}
	P_t \left(y_t \vert y_{<t},R\right) = \text{Softmax} \left(D_t W_p\right),
\label{eq6}
\end{equation}
\end{small}
where $R$ denotes the encoded video features, $W_p \in \mathbb{R}^{d_h \times d_e}$ is a trainable variable. The objective is to minimize the cross-entropy loss:
\begin{small}
\begin{equation}
	\mathcal{L}_t = -\sum^T_{t = 1} \log P_t \left(y^*_t \vert y^*_{<t}, R\right),
\end{equation}
\end{small}
where $y^*_t$ denotes the ground-truth token at time step $t$.

\subsection{Frequency-Aware Diffusion}
ORG-TRL~\cite{ZhangSY0WHZ20} introduces an external language model to extend the ground-truth tokens to cope with the token imbalance problem. 
However, it needs to be trained in large-scale corpus in advance, divorced from the captioning model, and the external model consumes additional computing resources and memory. 
Instead, we propose an FAD module inside captioning model to sufficiently grasp low-frequency semantics. 
We first introduce our split method of high-frequency and low-frequency tokens and present a reasonable explanation for it. Then we elaborate on the diffusion process to learn the semantics of low-frequency tokens.

\subsubsection{Split of Distinct Frequency Words.}
Due to the severe imbalance of tokens in the corpus, a small number of high-frequency tokens occupy many occurrences while a large number of low-frequency tokens appear rarely. Here we indicate HFT for high-frequency tokens, LFT for low-frequency ones, and UMT for unmarked tokens. Then we define a token belongs to different categories of frequency as:
\begin{small}
\begin{equation}
	\left\{
	\begin{aligned}
	\text{HFT}, \frac{\left|\text{vid} \left(\text{tok}\right)\right|}{\left|\text{vid}_\text{all} \left(\text{tok}\right)\right|}\geq\gamma\\
	\text{LFT}, \frac{\left|\text{vid} \left(\text{tok}\right)\right|}{\left|\text{vid}_\text{all} \left(\text{tok}\right)\right|}<\gamma
\end{aligned}
\right., \quad \frac{\left|\text{tok}\right|}{\left|\text{cap}\right|}\geq\delta,
\end{equation}
\begin{equation}
	\left\{
	\begin{aligned}
	\text{UMT}, \frac{\left|\text{vid} \left(\text{tok}\right)\right|}{\left|\text{vid}_\text{all} \left(\text{tok}\right)\right|}\geq\gamma\\
	\text{LFT}, \frac{\left|\text{vid} \left(\text{tok}\right)\right|}{\left|\text{vid}_\text{all} \left(\text{tok}\right)\right|}<\gamma
\end{aligned}
\right., \quad \frac{\left|\text{tok}\right|}{\left|\text{cap}\right|}<\delta,
\end{equation}
\end{small}
where $\delta$ and $\gamma$ represent hyper-parameter to decide which category of the frequency the token belongs to. 
In video captioning, a video has multiple ground-truth captions. $|\text{vid}(\cdot)|$ and $|\text{vid}_\text{all}(\cdot)|$ respectively represent the occurrence numbers of the token in the corresponding video and the sum number of tokens in all captions of its video.
$|\text{tok}|$ denotes the number of occurrences of this token in all ground-truth captions of the whole corpus, and $|\text{cap}|$ indicates the number of all captions in the dataset.

We use $\gamma$ to assess the frequency of the token in intra-video, for it only indicates the frequency inside its located video (termed intra-frequency), and $\delta$ evaluates the frequency of the token in inter-video for it expresses the frequency in the whole dataset (termed inter-frequency). We split tokens into three kinds of frequency tokens in four cases:
When inter-frequency exceeds the $\delta$, if the token has higher intra-frequency, it is divided into HFT owing to its universality; Otherwise, proving that it cannot integrate sufficiently with the corresponding video and be divided into LFT, even if it is a common token; 
When inter-frequency is below the $\delta$ while the intra-frequency is greater than the $\gamma$, it shows that even though it is not a frequent token, it can be adequately comprehended by the video to which it belongs. Thus it is divided into UMT for it can be self-sufficient without disturbing other high-frequency words; Instead, when both of them are below their respective hyper-parameters, the token is obviously classified as LFT, for it requires more trains.

\subsubsection{Noising Frequency Diffusion.}
Our diffusion is an intuitive diffusion exploiting one noise addition step rather than the vanilla Markov chain with multiple adding noise steps. Inspired by denoising after adding noise~\cite{HoJA20}, we regard low-frequency tokens as noise and add them to high-frequency tokens so that our diffusion is a process from low-frequency to high-frequency.
We respectively define $\text{Token}^L = \{\text{token}_1^L, \text{token}_2^L, \cdots, \text{token}_m^L\}$, $\text{Token}^H = \{\text{token}_1^H, \text{token}_2^H, \cdots, \text{token}_n^H\}$ as low-frequency and high-frequency token features list.
A similarity matrix $S$ is constructed to measure the similarity semantics between $i$-th LFT and $j$-th HFT:
\begin{small}
\begin{equation}
	S_{ij} = \frac{\exp s_{ij}}{\sum\nolimits_{j=1}^n \exp s_{ij}},
\end{equation}
\end{small}
where $S_{ij}$ is the $(i,j)$-th element of $S\in\mathbb{R}^{m \times n}$. We adopt softmax function as the normalization function, $s_{ij}$ denotes the cosine similarity score between two tokens. We choose $j$-th HFT corresponding to max value in each line as the diffusion object of $i$-th LFT:
\begin{small}
\begin{equation}
	\text{LoH} = \arg \max \left(S, \text{dim} = 1\right),
\end{equation}
\end{small}
where LoH denotes the list of high-frequency tokens corresponding to low-frequency tokens one by one, and the value of LoH represents the index of $\text{Token}^H$. In this way, we have established the corresponding relationship between low-frequency tokens and their semantically most similar high-frequency token. 
Afterward, we obtain the parameter of the noise according to the similarity:
\begin{small}
\begin{equation}
	\alpha_i = \frac{\exp \sigma \left(\text{token}^L_i, \text{token}^H_{\text{loh}}\right)}{\sum_{i \in N} \exp \sigma \left(\text{token}^L_i, \text{token}^H_{\text{loh}}\right)},
\end{equation}
\end{small}
where $\text{loh}$ denotes $i$-th value of $\text{LoH}$, $\alpha_i$ represents the significance of $i$-th noise, $N$ denotes the indexes of low-frequency words corresponding to one high-frequency word, $\sigma$ denotes the cosine similarity of semantic between these two tokens, then we obtain the diffusion $\text{token}^H_{\text{loh}}$ of high-frequency:
\begin{small}
\begin{equation}
	\text{token}^H_{\text{loh}} = \text{token}^H_{\text{loh}}+\sum_{i\in N} \alpha_i \circ \text{token}^L_i.
\end{equation}
\end{small}
Note that unlike DDPM~\cite{HoJA20} and TimeGrad~\cite{RasulSSV21}, which learns the noise first and then denoising, our model only learns low-frequency tokens regardless of how to remove them. 
By adding appropriate low-frequency tokens as noise to high-frequency ones, RSFD can generate relatively rarely used but more refined words.

\subsection{Divergent Semantic Supervisor}
Due to the noise semantics of low-frequency in FAD, the semantics of high-frequency is constrained to some extent. To complement the semantics of the central high-frequency tokens and to further encourage the formation of the central low-frequency tokens, DSS offers the contextual cue of adjacent tokens to the central tokens.


Skip-gram~\cite{word2vec} designs a window centered on the central word and supervises the generation of the central word using the contextual words. Motivated by it, we design a DSS obtaining the gradient that adapts to the token itself to understand the word adequately. 

As shown in Fig.~\ref{fig2}, we project the hidden features to the new feature space before the last linear layer and then project new features to the space of the dimension of corpus size:
\begin{small}
\begin{equation}
	D_t^{'} = D_t W_a W_p,
\end{equation}
\begin{equation}
	P_t^{'} \left(y_t \vert y_{<t}, R\right) = \text{Softmax} \left(D_t^{'}\right),
\end{equation}
\end{small}
where $D_t$ is the result of Eq.~\eqref{eq5}, $W_a \in \mathbb{R}^{d_h \times d_h}$ is a trainable variable, $W_p \in \mathbb{R}^{d_h \times d_e}$ is also a trainable variable shared with Eq.~\eqref{eq6}. We use $D_\text{tf}^{'}$ and $D_\text{tl}^{'}$ to respectively represent the former and the latter linear layer of the same layer as the original last union linear projection:
\begin{small}
\begin{align} 
	P_\text{tf}^{'} \left(y_t \vert y_{<t-1},R\right) & = \text{Softmax} \left(D_\text{tf}^{'}\right),\\
	P_\text{tl}^{'} \left(y_t \vert y_{<t+1},R\right) & = \text{Softmax} \left(D_\text{tl}^{'}\right).
\end{align}
\end{small}
The divergent loss is calculated as:
\begin{small}
\begin{equation}
	\mathcal{L}_\text{div} = -\sum^T_{t = 1}\log P^{'}_\text{tf}(y^*_t \vert y^*_{<t-1},R) - \sum^T_{t = 1}\log P^{'}_\text{tl}(y^*_t \vert y^*_{<t+1},R),
\end{equation}
\end{small}
where two items denote the former and the latter divergent losses, respectively. The final loss is calculated as:
\begin{small}
\begin{equation}
	\mathcal{L} = \mathcal{L}_t + \lambda\mathcal{L}_\text{div},
\label{eq19}
\end{equation}
\end{small}
where $\lambda$ is a variable parameter to decide the significance of $\mathcal{L}_\text{div}$. In the above example, we only use one word before and after to supervise the central word. In the experimental results section, we display the number of words selected as the supervisor to achieve the best generation effect.

\section{Experimental Results}

\subsection{Datasets and Evaluation Metrics}



MSR-VTT~\cite{XuMYR16} consists of 10,000 video clips, each annotated with 20 English captions and a category. MSVD~\cite{ChenD11} is a widely-used benchmark video captioning dataset collected from YouTube, composed of 1,970 video clips and roughly 80,000 English sentences. Following the standard split, we use the same setup as previous works~\cite{PanCHLGAN20,ye2022hierarchical}, which takes 6,513 video clips for training, 497 video clips for validation, and 2,990 video clips for testing on MSR-VTT, as well as 1,200, 100, and 670 videos for training, validation, and testing on MSVD.
The vocabulary size of MSR-VTT is 10,547, whereas that of MSVD is 9,468.

We present the results on four commonly used metrics to evaluate caption generation quality: BLEU-4 (B-4)~\cite{PapineniRWZ02}, METEOR (M)~\cite{BanerjeeL05}, ROUGE\_L (R)~\cite{lin2004rouge}, and CIDEr (C)~\cite{VedantamZP15}. 
BLEU-4, METEOR, and ROUGE\_L are generally used in machine translation. CIDEr is proposed for captioning tasks specifically and is considered more consistent with human judgment. 

\subsection{Implementation Details}

In our experiments, we follow~\cite{PeiZWKST19} to extract image and motion features to encode video information. Specifically, we use ImageNet~\cite{DengDSLL009} pre-trained ResNet-101~\cite{HeZRS16} to extract 2D scene features for each frame. We also utilize Kinetics~\cite{kay2017Kinetics} pre-trained ResNeXt-101 with 3D convolutions~\cite{HaraKS18}.

In our implementation, size of video feature $d_v$ and the hidden size $d_h$ are set to 2,048 and 512. Empirically, we set the sampled frames $K=8$ for each video clip. We set maximum sequence length $T$ to 30 on MSR-VTT, whereas $T=20$ on MSVD. Transformer decoder has a decoder layer, 8 attention heads, 0.5 dropout ratio, and 0.0005 $\ell2$ weight decay. 
We implement word embeddings by trainable 512 dimensions of embedding layers. 
In the training phase, we adopt Adam~\cite{KingmaB14} with an initial learning rate of 0.005 to optimize our model. The batch size is set to 64, and the training epoch is set to 50. During testing, we use the beam-search method with size 5 to generate the predicted sentences. 
$\gamma$ and $\delta$ for deciding the category of the token are respectively set to 0.015 and 0.0015. We set $\lambda = 0.07$ on MSR-VTT and $\lambda = 0.4$ on MSVD to demonstrate the significance of the divergent loss. 
All our experiments are conducted on two NVIDIA Tesla PH402 SKU 200.

\begin{table*}
 	\small
	\centering
	\begin{tabular}{lccccccccc}
	\toprule
	\multirow{2}[2]{*}{Method} & \multirow{2}[2]{*}{Venue} & \multicolumn{4}{c}{MSR-VTT} & \multicolumn{4}{c}{MSVD}\\
	\cmidrule(lr){3-6}\cmidrule(lr){7-10} 
	& & B-4 & M & R & C & B-4 & M & R & C\\ 
	\midrule 
	M3-VC~\cite{Wang000T18} & CVPR '18 & 38.1 & 26.6 & - & - & 51.8 & 32.5 & - & -\\ 
 	RecNet~\cite{Wang00018} & CVPR '18 & 39.1 & 26.6 & 59.3 & 42.7 & 52.3 & 34.1 & 69.8 & 80.3\\
 	PickNet~\cite{ChenWZH18} & ECCV '18 & 41.3 & 27.7 & 59.8 & 44.1 & 52.3 & 33.3 & 69.6 & 76.5\\
	OA-BTG~\cite{ZhangP19a} & CVPR '19 & 41.4 & 28.2 & - & 46.9 & \textbf{56.9} & {36.2} & - & 90.6\\
	MARN~\cite{PeiZWKST19} & CVPR '19 & 40.4 & 28.1 & 60.7 & 47.1 & 48.6 & 35.1 & 71.9 & 92.2\\
	MGSA~\cite{ChenJ19} & AAAI '19 & 42.4 & 27.6 & - & 47.5 & 53.4 & 35.0 & - & 86.7\\
	GRU-EVE~\cite{AafaqALGM19} & CVPR '19 & 38.3 & 28.4 & 60.7 & 48.1 & 47.9 & 35.0 & 71.5 & 78.1\\
	POS-CG~\cite{Wang00JWL19} & ICCV '19 & 42.0 & 28.2 & 61.6 & 48.7 & 52.5 & 34.1 & 71.3 & 88.7\\
	STG-KD~\cite{PanCHLGAN20} & CVPR '20 & 40.5 & 28.3 & 60.9 & 47.1 & 52.2 & \textbf{36.9} & \textbf{73.9} & 93.0\\
	SAAT~\cite{ZhengWT20} & CVPR '20 & 40.5 & 28.2 & 60.9 & 49.1 & 46.5 & 33.5 & 69.4 & 81.0\\
	ORG-TRL~\cite{ZhangSY0WHZ20} & CVPR '20 & \textbf{43.6} & 28.8 & {62.1} & 50.9 & 54.3 & {36.4} & \textbf{73.9} & {95.2}\\
	SBAT~\cite{JinHCLZ20} & IJCAI '20 & 42.9 & {28.9} & 61.5 & {51.6} & 53.1 & 35.3 & 72.3 & 89.5\\
	TTA~\cite{TuZGGY21} & PR '21 & 41.4 & 27.7 & 61.1 & 46.7 & 51.8 & 35.5 & 72.4 & 87.7\\
	SibNet~\cite{LiuRY21} & TPAMI '21 & 41.2 & 27.8 & 60.8 & 48.6 & {55.7} & 35.5 & 72.6 & 88.8\\
	SGN~\cite{RyuKKY21} & AAAI '21 & 40.8 & 28.3 & 60.8 & 49.5 & 52.8 & 35.5 & {72.9} & 94.3\\
	AR-B~\cite{YangZLZ21} (Baseline) $\dag$ & AAAI '21 & 42.1 & {29.1} & 61.2 & 51.1 & 49.2 & 35.3 & 72.1 & 91.4\\
	FrameSel~\cite{LiZTXLT22} & TCSVT '22 & 38.4 & 27.2 & 59.7 & 44.1 & 50.4 & 34.2 & 70.4 & 73.7\\
	TVRD~\cite{WuNYXZW22} & TCSVT '22 & 43.0 & 28.7 & 62.2 & 51.8 & 50.5 & 34.5 & 71.7 & 84.3\\
	\midrule
	RSFD (Ours) & & {43.4} & \textbf{29.3} & \textbf{62.3} & \textbf{53.1} & 51.2 & 35.7 & {72.9} & \textbf{96.7}\\
	\bottomrule
	\end{tabular}
	\caption{Performance (\%) comparison with the state-of-the-arts on MSR-VTT and MSVD. $\dag$ indicates the reproduced method. The best results are shown in bold.}
	\label{table1}
\end{table*}

\subsection{Comparison with the State-of-the-Arts}
The comparison results with the state-of-the-arts in Table~\ref{table1} illustrate that RSFD achieves a significant performance improvement. On MSR-VTT, RSFD achieves the best results under 3 out of 4 metrics. 
Specifically, RSFD outperforms existing methods except for ORG-TRL~\cite{ZhangSY0WHZ20} under all the metrics on MSR-VTT, proving that our focus on solving the insufficient occurrence of low-frequency words is practical. 
In particular, RSFD achieves 53.1\% and 96.7\% under CIDEr on MSR-VTT and MSVD, respectively, making an improvement of 2.0\% and 5.3\% over AR-B~\cite{YangZLZ21}. 
As CIDEr captures human judgment of consensus better than other metrics, superior performance under CIDEr indicates that RSFD can generate semantically more similar to human understanding. 
The boost in performance demonstrates the advantages of RSFD, which exploits FAD and DSS, and models the low-frequency token for detailed and comprehensive textual representations.

RSFD is slightly lower than ORG-TRL under BLEU-4 on MSR-VTT, which uses an external language model to integrate linguistic knowledge.
But it is undeniable that RSFD is second only to it, demonstrating the superiority of RSFD.
OA-BTG~\cite{ZhangP19a}, STG-KD~\cite{PanCHLGAN20}, and ORG-TRL perform better under METEOR and ROUGE\_L on MSVD.
They either capture the temporal interaction of objects well in a small dataset or bring complementary representation by considering video semantics. 
Despite not exploring the above information, RSFD focuses on generating refined descriptions and outperforms other methods under CIDEr, benefiting from its emphasis on refined low-frequency tokens. Indicated that CIDEr best matches human consensus judgment.


\begin{table}
 	\small
	\centering
	\begin{tabular}{cccccc}
	\toprule
	Dataset & size & B-4 & M & R & C\\ 
	\midrule
	\multirow{4}{*}{MSR-VTT} & 3 & 42.5 & 29.2 & 61.6 & 52.2\\
	& 5 & \textbf{43.4} & \textbf{29.3} & \textbf{62.3} & \textbf{53.1}\\
	& 7 & 42.9 & 29.2 & 61.5 & 51.9\\
	& 9 & 42.7 & 29.1 & 61.4 & 51.6\\
	\midrule
	\multirow{4}{*}{MSVD} & 3 & 50.9 & 35.3 & 72.5 & 94.1\\
	& 5 & \textbf{51.2} & \textbf{35.7} & \textbf{72.9} & \textbf{96.7}\\
	& 7 & 50.6 & 35.4 & 72.6 & 93.8\\
	& 9 & 50.7 & 35.3 & 72.4 & 93.0\\
	\bottomrule
	\end{tabular}
	\caption{Performance (\%) comparison of different configurations of window size in DSS module on MSR-VTT and MSVD. The best results are shown in bold.}
	\label{table2}
\end{table}

\subsection{Study on Trade-off Parameter}
\subsubsection{Trade-off Parameter of Supervision Window.}
To explore the impact of the number of words, we design a supervision window of different sizes to extract adjacent supervision information shown in Table~\ref{table2}. It shows that when the window size increases, the overall performance shows an upward trend, but when the size exceeds 5, the performance begins to decline.
When the supervision window is too small, insufficient adjacent information cannot play a supervisory role. And in an oversized supervision window, not only low-frequency tokens, all tokens receive unique gradient updates, causing overfitting of other tokens that have been trained enough.


\begin{figure*}
	\centering
	\includegraphics[width = \textwidth]{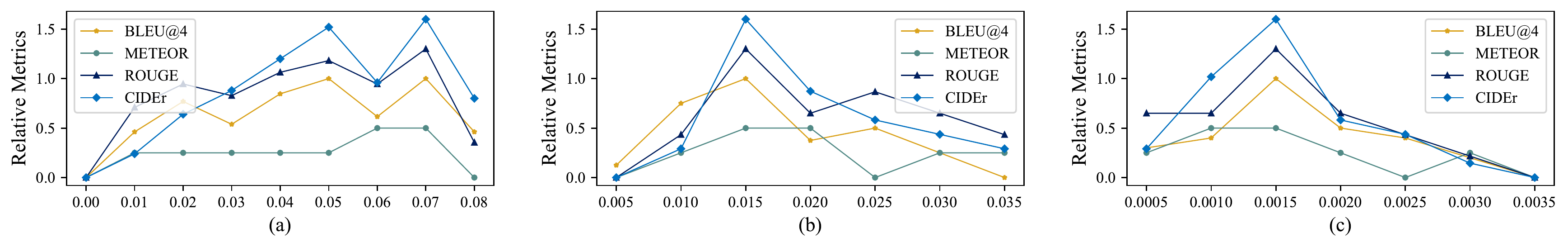}
	\caption{Analysis of (a) different parameter $\lambda$ for the significance of divergent semantic supervisor, (b) different parameter $\gamma$ deciding intra-frequency, and (c) different parameter $\delta$ determining inter-frequency on MSR-VTT. We show the relative results on all metrics.}
	\label{fig3}
\end{figure*}

\subsubsection{Trade-off Hyper-Parameter $\lambda$.}
To evaluate the effectiveness of $\lambda$ and ﬁnd an appropriate value for $\lambda$, we adjust the value of Eq.~\eqref{eq19} under all metrics on MSR-VTT are given in Fig.~\ref{fig3}(a). Note that only using original loss ($\lambda = 0$) shows the worst performances. 
Thus it can be concluded again that adding divergent loss boosts the performance of the captioning model, which benefited from gradient correction of adjacent words to central words.
Considering all the metrics comprehensively, we empirically set $\lambda = 0.07$ on MSR-VTT.

\subsubsection{Trade-off Hyper-Parameter $\gamma$ and $\delta$.}
We conduct experiments on MSR-VTT to explore the effect of hyper-parameter $\gamma$ for the intra-frequency and hyper-parameter $\delta$ for the inter-frequency. 
All metrics scores of different settings are illustrated in Figs.~\ref{fig3}(b) and \ref{fig3}(c). If $\delta$ is too high, the reduction of the number of high-frequency tokens makes the information of low-frequency tokens cannot be well absorbed and fully integrated with visual features. 
Similarly, if $\gamma$ is too high, some words that should not be split into low-frequency tokens are divided into them, causing the overfitting of these words. In contrast, if $\gamma$ is set too low, low-frequency tokens that need to be sufficiently trained are ignored. Considering the above discussion, we set $\gamma$ and $\delta$ as 0.015 and 0.0015, respectively.

\begin{table}
 	\small
	\centering
	\begin{tabular}{clcccc}
	\toprule
	Dataset & Method & B-4 & M & R & C\\ 
	\midrule
	\multirow{4}{*}{MSR-VTT} & AR-B (Baseline) & 42.1 & 29.1 & 61.2 & 51.1\\
	& RSFD w FAD & 43.0 & 29.2 & 62.0 & 52.9\\
	& RSFD w DSS & 42.5 & 29.2 & 61.6 & 52.2\\
	& RSFD (Ours) & \textbf{43.4} & \textbf{29.3} & \textbf{62.3} & \textbf{53.1}\\
	\midrule
	\multirow{4}{*}{MSVD} & AR-B (Baseline) & 49.2 & 35.3 & 72.1 & 91.4\\
	& RSFD w FAD & 50.9 & 35.5 & 72.7 & 94.1\\
	& RSFD w DSS & 50.6 & 35.5 & 72.3 & 93.1\\
	& RSFD (Ours) & \textbf{51.2} & \textbf{35.7} & \textbf{72.9} & \textbf{96.7}\\
	\bottomrule
	\end{tabular}
	\caption{Performance (\%) comparison of different components in RSFD on MSR-VTT and MSVD. The best results are shown in bold.}
	\label{table3}
\end{table}

\subsection{Ablation Study}
\subsubsection{Effectiveness of Different Components.}
The results are presented in Table~\ref{table3}. 
The first line method AR-B~\cite{YangZLZ21} (baseline) only applies a Transformer-based model to the encoder-decoder framework. The methods in the second and third rows refer to methods with FAD and DSS, respectively. 
It can be observed that FAD, through sufficiently exploiting the semantics of low-frequency tokens, outperforms AR-B under popular metrics, particularly bringing significant CIDEr improvements. Although FAD slightly affects the semantics of high-frequency tokens, we integrate it with DSS, which makes up the information of high-frequency tokens to some extent. Our complete method RSFD results demonstrate that combining the above two modules can achieve the best performance.

\begin{figure}
	\centering
	\includegraphics[width = \columnwidth]{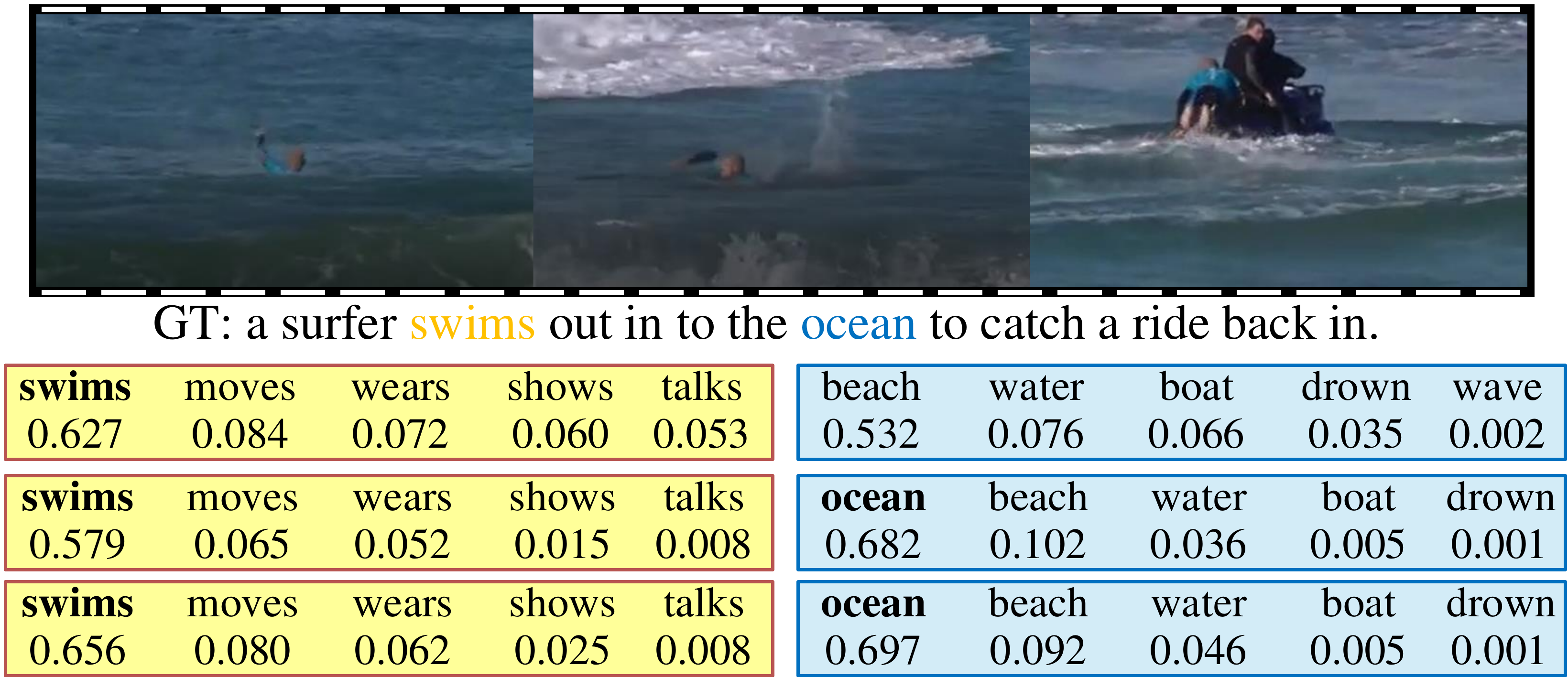} 
	\caption{Instances of AR-B (baseline), RSFD with FAD, and RSFD model in the testing phase. The yellow and blue words indicate the high-frequency and low-frequency words to be predicted. The yellow box on the left corresponds to the three methods predicting the yellow word, while the blue box on the right generates the blue word.}
	\label{fig4}
\end{figure}

\subsubsection{Evaluation of FAD and DSS.}
We analyze the effect of each component in the testing phase shown in Fig.~\ref{fig4}. 
It can be seen that FAD generates ``ocean'' shown in the second line, which as an infrequent token, carries critical semantics but is challenging to predict by AR-B~\cite{YangZLZ21}. FAD helps the model comprehend low-frequency tokens relative to video content. 
Although FAD slightly narrows the high-frequency token “swims” probability from 0.627 to 0.579, DSS compensates for it to 0.656 and further enhances the semantics of the low-frequency token “ocean” from 0.682 to 0.697.
In general, our method promotes the generation of refined captions owing to our focus on low-frequency tokens and effectively alleviates the long-tailed problem in video captioning, demonstrating a competitive generation effect.

\begin{figure}
	\centering
	\includegraphics[width = \columnwidth]{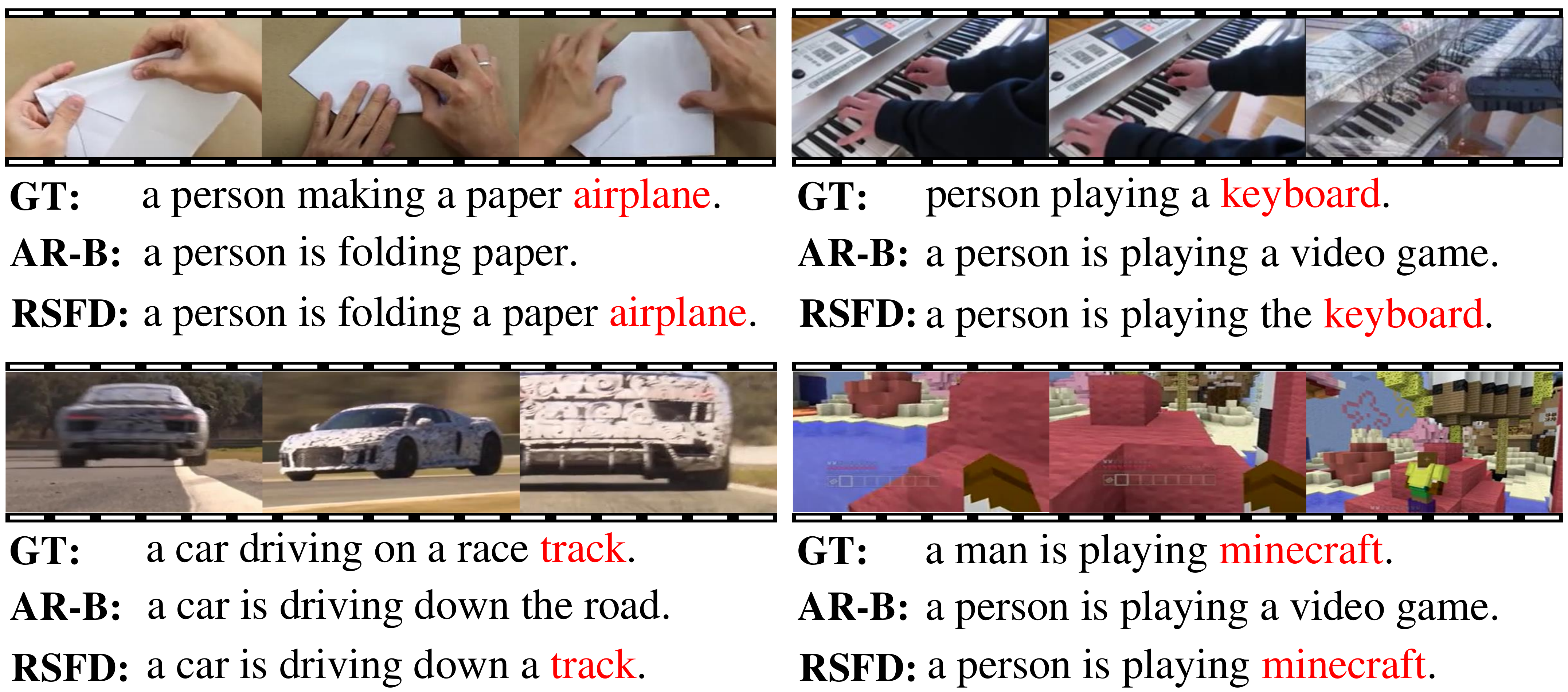} 
	\caption{Qualitative results generated on the testing sets of MSR-VTT with AR-B (baseline) and RSFD. The red words indicate the refined information in the ground truth as well as the generated sentences of RSFD.}
	\label{fig5}
\end{figure}

\subsection{Qualitative Results}
We present some results generated on MSR-VTT in Fig.~\ref{fig5}. It can be observed that the content of captions generated by RSFD is more refined than AR-B~\cite{YangZLZ21} (baseline). Taking the first result as an example, AR-B only understands the rough meaning of the video, \textit{e.g.}, ``folding'', missing the abundant description, \textit{e.g.}, ``airplane''. By contrast, our model captures more details and generates them accurately, demonstrating that RSFD produces relatively less common but critical words. The rest of the results have similar characteristics.

\section{Conclusion}
In this paper, we present a Refined Semantic enhancement towards Frequency Diffusion (RSFD) for video captioning.
RSFD addresses the problem of the Transformer-based or RNN-based architecture on the insufficient occurrence of low-frequency tokens that limits the capability of comprehensive semantic generation. Our method incorporates the Frequency-Aware Diffusion (FAD) and Divergent Semantic Supervisor (DSS), which capture more detailed semantics of relatively less common but critical tokens to provide refined semantic enhancement with the help of low-frequency tokens. 
RSFD achieves competitive performances by outperforming baselines by large margins of 2.0\% and 5.3\% in terms of CIDEr on MSR-VTT and MSVD, respectively.


\section*{Acknowledgments}
This work was supported in part by the National Natural Science Foundation of China under Grants 62271361 and 62066021, the Department of Science and Technology, Hubei Provincial People’s Government under Grant 2021CFB513, and the Hubei Province Educational Science Planning Special Funding Project under Grant 2022ZA41.

\bibliography{RSFD}

\end{document}